\documentclass[10pt,twocolumn,letterpaper]{article}

\usepackage{icb}
\usepackage{times}
\usepackage{epsfig}
\usepackage{graphicx}
\usepackage{amsmath}
\usepackage{amssymb}
\usepackage{gensymb}
\usepackage{multirow}

\usepackage[norule,symbol,perpage]{footmisc}

\usepackage[pagebackref=true,breaklinks=true,letterpaper=true,colorlinks,bookmarks=false]{hyperref}

\icbfinalcopy 

\ificbfinal\fi

\begin{document}

\title{Domain Adaptation in Multi-Channel Autoencoder based Features for Robust Face Anti-Spoofing}

\author{Olegs Nikisins, Anjith George, S\'ebastien Marcel \\
Idiap Research Institute \\
Rue Marconi 19, CH - 1920, Martigny, Switzerland \\
{\tt\small  \{olegs.nikisins, anjith.george, sebastien.marcel\}@idiap.ch  }
}

\maketitle
\thispagestyle{empty}

\begin{abstract}

   While the performance of face recognition systems has improved significantly in the last decade, they are proved to be highly vulnerable to presentation attacks (spoofing). Most of the research in the field of face presentation attack detection (PAD), was focused on boosting the performance of the systems within a single database. Face PAD datasets are usually captured with RGB cameras, and have very limited number of both bona-fide samples and presentation attack instruments. Training face PAD systems on such data leads to poor performance, even in the closed-set scenario, especially when sophisticated attacks are involved.
   We explore two paths to boost the performance of the face PAD system against challenging attacks. First, by using multi-channel (RGB, Depth and NIR) data, which is still easily accessible in a number of mass production devices. Second, we develop a novel Autoencoders + MLP based face PAD algorithm. Moreover, instead of collecting more data for training of the proposed deep architecture, the domain adaptation technique is proposed, transferring the knowledge of facial appearance from RGB to multi-channel domain. We also demonstrate, that learning the features of individual facial regions, is more discriminative than the features learned from an entire face. The proposed system is tested on a very recent publicly available multi-channel PAD database with a wide variety of presentation attacks.

\end{abstract}

\section{Introduction}
\label{sec:introduction}

Presentation Attacks Detection (PAD), also known as anti-spoofing, has gained a significant attention in the biometric society, since high accuracy state-of-the-art face recognition methods are known to be vulnerable to presentation attacks~\cite{Bhattacharjee_BTAS2018_2018, Mohammadi_IETBIOMETRICS_2017}.
This loophole in the security of recognition and verification systems is unacceptable in high security applications, such as border control or law enforcement, dictating the need of having a human-in-the-loop co-supervising the recognition process. Also, the fabrication of Presentation Attack Instruments (PAI) is getting more trivial, due to common availability of dozens of biometric samples in social networks, and improving fabrication technologies, such as 3D printers.
Thus, the availability of \textit{high accuracy} face PAD systems is the missing component in the wide deployment of face recognition technologies.

Presentation attacks in general are of unknown nature, meaning that it can be anything as long as it helps the attacker either \textit{impersonate} or \textit{obfuscate} the identity. Despite this fact, a vast majority of the research in face PAD, focuses on two types of attacks: photo and replay PAI. One explanation to this phenomena is a domination of publicly available databases containing just these two types of attacks, for example SiW~\cite{siw_db}, Replay-Mobile~\cite{Costa-Pazo_BIOSIG2016_2016}, OULU-NPU~\cite{OULU_NPU_2017}, MSU MFSD~\cite{WenTIFS15}, or aggregations of such~\cite{Nikisins_anomaly_PAD}. Also, the biometric samples in these databases are RGB only. Recent research~\cite{Nikisins_anomaly_PAD, Kittler_anomaly_PAD} states that RGB-based face PAD systems have relatively low performance in general, even for fore-mentioned two PAI types, and the situation is getting worse in the unseen-attacks tests. In our experiments we use a database containing a much wider range of PAIs, both 2D and 3D, as well as partial attacks, demonstrating that RGB-based PAD performs poorly even in the seen-attacks scenario, which coincides with~\cite{Nikisins_anomaly_PAD, Kittler_anomaly_PAD}. The solution boosting the performance of PAD system is to use multi-channel (\textbf{MC}) based approach, enhancing PAD system with specialized imaging sensors. Most of the effort with specialized sensors for facial biometrics was concentrated on face recognition (\textbf{FR})~\cite{7746050}. However, recently some authors pointed out the applicability of this idea to face PAD. A preliminary study using NIR and LWIR and Depth (\textbf{D}) cameras for face PAD was introduced in ~\cite{8053524}. In ~\cite{8053524} authors argue that not just simple PAI, \eg photo and replay PAI, but also advanced attacks, such as silicon masks, should be detectable much easier using MC PAD approach, however they don't introduce any practical algorithmical solution in the paper. In~\cite{DBLP:journals/js/SteinerSKJ16} a SWIR-camera based skin detection methodology is introduced, being potentially applicable to face PAD task, assuming that in the attack attempts the large fraction of the face is covered with synthetic materials. The potential challenge for skin-detection based PAD is partial attacks, covering just a small fraction of the face potentially important for FR methods, \eg eye region.
In~\cite{8009749} authors developed a PAD method using multi-spectral camera, capturing the samples in 7 spectral bands from 425 to 930 (nm). They proposed an algorithm based on hand-crafted, LBP and BSIF, features and SVM classifier, as well as various data fusion strategies at image and score levels. The reported results are promising, however the PAI types used in the experiments of~\cite{8009749} are limited to photo attacks only. To the best of our knowledge, our work is the first attempt introducing deep-learning based multi-channel face PAD system efficiently detecting a wide range of PAIs, such as 2D, 3D, and partial attacks.

The core of the \textit{proposed} \textbf{MC face PAD} algorithm is a Convolutional Neural Network (CNN), which is decomposed into two components: a \textbf{set of MC encoders} extracted from pre-trained autoencoders (\textbf{AE}), and a \textbf{Multi-Layer Perceptron} (\textbf{MLP}) combining previous set of encoders. A task of the \textbf{set of MC encoders} is \textit{feature extraction} from multi-channel input data, which in our case is a stack of gray-scale, NIR, and Depth (\textbf{BW-NIR-D}) facial images, or a stack of images of facial features, \eg left-eye region. Is remarkable, that AE are trained using bona-fide samples only, learning the appearance of the real face. Moreover, instead of collecting more data for AE training, the domain adaptation technique is proposed, transferring the knowledge of facial appearance from RGB to BW-NIR-D domain. A task of \textbf{MLP} is \textit{classification} categorizing the features of MC encoders as either bona-fide or an attack. Only MLP is trained using samples from both bona-fide and attack classes.
To the best of our knowledge, both proposed Autoencoder+MLP based PAD algorithm, and domain adaptation (transfer learning) approach, are unique in the field of MC face PAD.
Most of the research in the field of face PAD rely on the binary classifiers, \eg SVM or LR, categorizing hand-crafted features, such as LBP or IQM~\cite{Ramachandra:2017:PAD:3058791.3038924}. In~\cite{8009749} similar strategy is applied to the task of MC face PAD, also testing different strategies for channels fusion. Relatively recent trend in face PAD is to use deep-learning based techniques.
In~\cite{7821013} authors are using transfer learning ideas fine-tuning VGG-Face model~\cite{Parkhi15}, CNN originally designed for face recognition, to the face spoofing datasets. Then PCA is used reducing the dimensionality of the feature vectors, which are next classified with SVM. The experiments are done on CASIA-FA~\cite{6199754}, Replay-Attack~\cite{Chingovska_BIOSIG-2012} containing photo and replay PAI only.
Another CNN-based face PAD paper~\cite{DBLP:journals/corr/abs-1806-07492} is focusing on initialization procedure of CNN weights, arguing it improves the convergence of the training and overall performance of the system. Authors~\cite{DBLP:journals/corr/abs-1806-07492} first train a set of 9 patch-CNNs, learning features of a different facial regions. Afterwards, weights of patch-CNNs are substituted into a single CNN, which is then fine-tuned on the whole face. Again, a set of PAI in experimental section of~\cite{DBLP:journals/corr/abs-1806-07492} is limited to photo and video attacks.
In~\cite{DBLP:conf/cvpr/LiuJ018} CNN-RNN model is proposed estimating two biometric traits from an input RGB video - depth pattern of the face, and rPPG signal. The results on SiW~\cite{siw_db}, and OULU-NPU~\cite{OULU_NPU_2017} are promising, however from problem formulation, one can conclude that partial attacks can be problematic for this approach, since both rPPG and shape are preserved in this case. The are also attempts to use autoencoders for face PAD, in~\cite{face-upad} authors combine hand-crafted LBP features and an autoencoder based outliers detection algorithm. It aims to detect unknown PAI better than other methods using hand-crafted features, however overall performance on OULU-NPU~\cite{OULU_NPU_2017} is relatively low.
In~\cite{DBLP:journals/corr/abs-1802-05798} authors suggest an interesting methodology on detecting anomalies in the face, \eg partial attacks, however they don't discuss the application of the algorithm to the face PAD. To the best of our knowledge, there is no previous published research using deep-learning based approaches for MC face PAD, making our paper an initial contribution to this promising direction.

To summarize, the following \textbf{main contributions} are proposed in our paper. \textit{First}, a novel deep-learning based MC face PAD algorithm is introduced, having a CNN-like structure composed of a \textit{set of MC encoders} and an \textit{MLP}. \textit{Second}, instead of collecting more training data, the domain adaptation technique is proposed, transferring the knowledge of facial appearance from RGB to multi-channel domain.
Domain adaptation is done via autoencoders, which are first pre-trained on a large publicly available RGB face database, and are then \textit{partially} fine-tuned on the set of BW-NIR-D face images. \textit{Third}, we demonstrate, that learning the features of individual facial regions, is more discriminative than the features learned from an entire face. Proposed MC face PAD method gives very promising results, significantly outperforming hand-crafted features based baseline, on the challenging database, namely WMCA, containing a wire rage of PAI - 3D, 2D, and partial attacks.
\textit{Finally}, the results reported in this work are fully reproducible: experimental database is publicly available, the evaluation protocols are strictly defined, and the source code for replicating experiments is published\footnote{Source code: \url{https://gitlab.idiap.ch/bob/bob.paper.mcae.icb2019}}.

\section{Proposed multi-channel face PAD approach}
\label{sec:proposed_multichannel_face_pad_approach}

\begin{figure*}[t]
\begin{center}
   \includegraphics[width=0.6\linewidth]{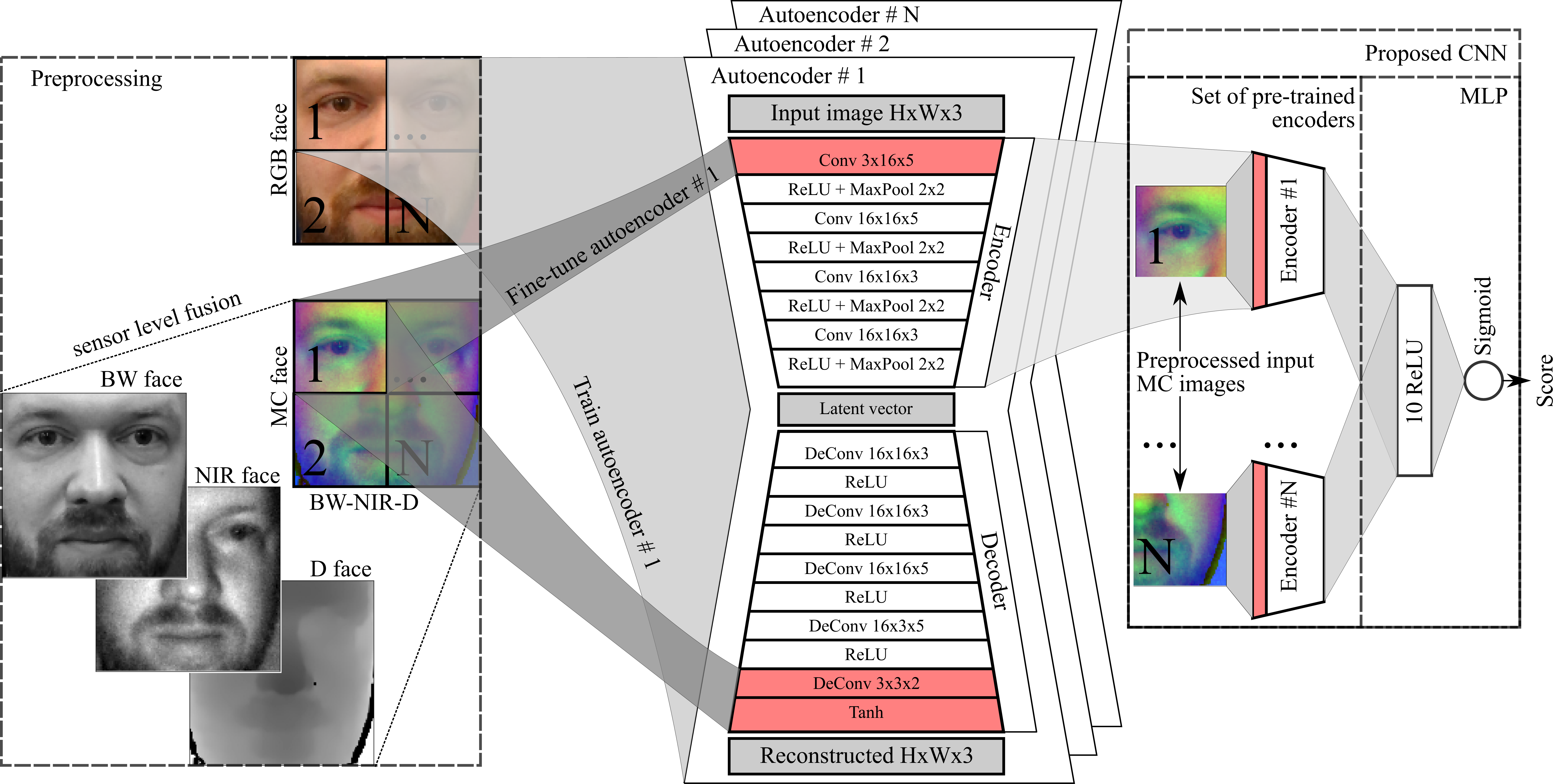}
\end{center}
   \caption{Schematic representation of the proposed MC face PAD approach: visualizing the preprocessing stage, internal structure of convolutional autoencoders, training and domain adaptation (fine-tuning) strategy, and the structure of the final CNN-based PAD system. The Conv/DeConv layers are parametrized as follows: number of input channels $\times$ number of output channels $\times$ size of the filter kernel.}
\label{fig:system_block_diagram}
\end{figure*}

This section briefly introduces the proposed multi-channel face PAD system. The PAD algorithms is CNN based, where special CNN structure is used to categorize input MC data into bona-fide or attack classes, see Figure~\ref{fig:system_block_diagram}.
While the network architecture can be represented as CNN, the proposed training technique, allowing domain adaptation, is unique, and is discussed later.
An input MC data for the CNN is a \textit{set of stacked BW-NIR-D} images corresponding to different facial regions, \eg left-eye, see Figure~\ref{fig:system_block_diagram}. According to the widely accepted taxonomy~\cite{IJST120575}, this type of biometric information integration from multiple cameras is called \textit{sensor level fusion}. Sensor level fusion preserves maximum variance in the fused data, allowing the network to assess the hidden dependencies in the lowest level. However, this type of fusion requires special data preprocessing, making the streams to be blended compatible.

\begin{table}[h]
\begin{center}
\footnotesize
\begin{tabular}{c|c|c}
\hline
\# & Input stream & Preprocessing \\ \hline\hline
1. & RGB     & Face, and landmarks detection in all frames.      \\ \hline
2. & RGB     & Conversion of all frames to BW format.            \\ \hline
3. & NIR, D  & Registration / alignment to BW channel.           \\ \hline
4. & NIR, D  & Normalization of dynamic range to $[0, 255]$.     \\ \hline
5. & NIR, D  & Data type-casting to 8-bit format.                \\ \hline
6. & BW, NIR, D & Scaling, rotation, and cropping of facial regions.  \\ \hline
7. & BW, NIR, D & Stacking of facial images into BW-NIR-D image.      \\ \hline
\end{tabular}
\end{center}
\caption{Preprocessing steps to generate BW-NIR-D facial images.}
\label{tab:preprocessing_steps}
\end{table}

\textbf{Preprocessing} applied to all frames of input RGB, NIR, and D videos, allowing to generate BW-NIR-D facial images, is summarized in Table~\ref{tab:preprocessing_steps}. An example of the preprocessing product is displayed in Figure~\ref{fig:system_block_diagram}, denoted as MC face, and RGB face. To generate RGB facial images, only steps 1 and 6 are applied to the input RGB stream.
The face and landmark detection method is using MTCNN~\cite{7553523} deep neural network.
The proposed \textit{dynamic range normalization} technique of the NIR, and D streams is based on Median Absolute Deviation (MAD) measure. To clarify the MAD-based normalization, let $\mathbf{I}$ be a non-RGB image of the facial region. Given $\mathbf{I}$, a vector $\mathbf{v}$ containing non-zero elements of $\mathbf{I}$ is obtained. Next, a MAD measure is computed as follows:
\begin{equation}
\label{eq:mad_calculation}
\mathbb{MAD} = median(|\mathbf{v}-median(\mathbf{v})|)
\end{equation}
Given the $\mathbb{MAD}$ value, the input image $\mathbf{I}$ is normalized:
\begin{equation}
\label{eq:mad_image_normalization}
\hat{\mathbf{I}}_{i,j} = \frac{(\mathbf{I}_{i,j} - median(\mathbf{v}) + \sigma \cdot \mathbb{MAD})}{2\cdot\sigma\cdot\mathbb{MAD}} \cdot (2^8-1),
\end{equation}
where $\hat{\mathbf{I}}$ is normalized image, $i = 1,\dots,W$; $j = 1,\dots,H$, and $W$, $H$ are the width and height of $\mathbf{I}$.
In our experiment: $\sigma_{NIR} = 3$, $\sigma_{D} = 6$.  The size of the facial images is normalized to $128 \times 128$ pixels, and the images are rotated so that the eye-line is horizontal. Given preprocessed training data the subsequent step is CNN training.

\begin{table}[h]
\begin{center}
\footnotesize
\begin{tabular}{c|c|c|c}
\hline
\# & Train step        & Training data        & DB, classes used                              \\ \hline\hline
1. & Train $N$ AEs     & RGB face regions     & CelebA~\cite{liu2015faceattributes}, BF       \\ \hline
2. & Fine-tune $N$ AEs & MC face regions      & WMCA, BF                                      \\ \hline
3. & Train an MLP      & MC latent encodings  & WMCA, BF\&PA                                  \\ \hline
\end{tabular}
\end{center}
\caption{Steps to train a CNN-based MC face PAD system. BF and PA stands for samples from \textit{bona-fide} and \textit{presentation attack} classes.}
\label{tab:training_steps}
\end{table}

\textbf{Training} of the proposed CNN based multi-channel face PAD system is using both RGB and MC facial data, and is summarized in Table~\ref{tab:training_steps}. An entire CNN is never trained, instead training steps are associated to two internal CNN components: a set of convolutional autoencoders, and an MLP, see Figure~\ref{fig:system_block_diagram}. First, a set of $N$ convolutional autoencoders, reconstructing RGB facial regions, is pre-trained. Pre-training is done using images from publicly available database, namely CelebA~\cite{liu2015faceattributes}, having significantly higher number of bona-fide samples and identities, than any PAD-specific database to date. AE pre-training helps to better position the network in the search space. The subsequent step is fine-tuning of AEs using MC face regions, extracted from train set of WMCA. It is worth mentioning, that the dimensionality of RGB and MC training samples is identical.
Here we propose to \textit{fine-tune just the first layers of encoders} (partial fine-tuning), and last 2 layers of decoders (for symmetry), instead of fine-tuning the entire autoencoders. The intuition behind this proposition, is that only \textit{low level features are domain dependent}, while deeper features are domain independent mostly preserving structural information of the face. In the experimental section we prove empirically, that proposed RGB-to-MC domain adaptation via \textit{partial fine-tuning} is more efficient than full fine-tuning.
The effectiveness of similar domain adaptation strategy has also been addressed in~\cite{deFreitasPereira_IEEET-IFS_2019}, focusing on the task of heterogeneous face recognition. Both pre-training and fine-tuning of AEs is using bona-fide samples only, and is taking 50 epochs each. The \textit{latent vectors} of trained encoders are used as input features for the MLP, see Figure~\ref{fig:system_block_diagram}. An MLP is trained using latent features of both bona-fide and attack classes present in the training set of the WMCA database, Table~\ref{tab:training_steps}. The best performing MLP model is selected cross-validating them on the development set of WMCA. Summarizing, the composition of trained convolutional autoencoders, and MLP, is forming the proposed MC face PAD system.

\section{Experimental Database}
\label{sec:experimental_database}

\begin{figure}[t]
\begin{center}
   \includegraphics[width=0.95\linewidth]{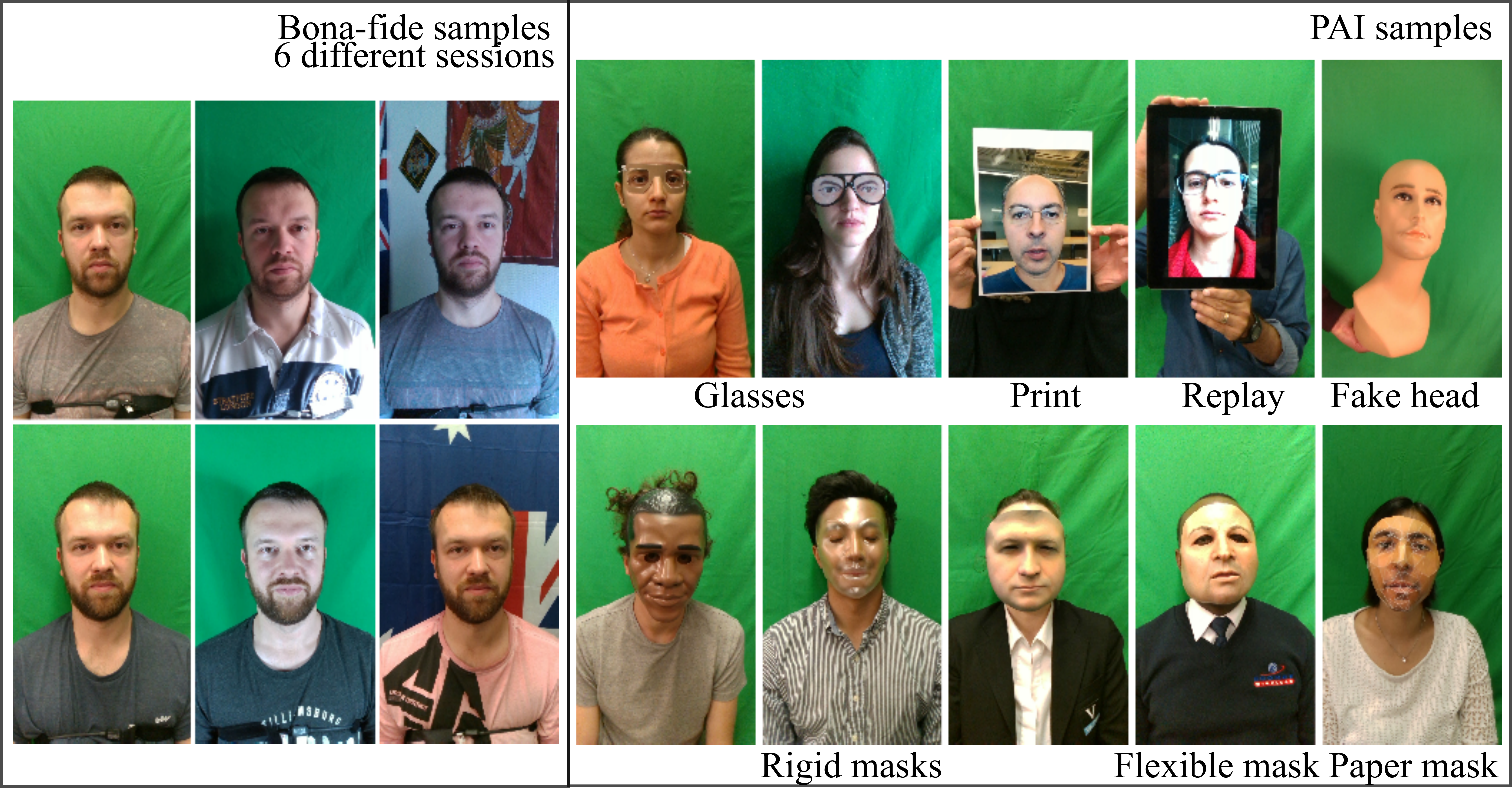}
\end{center}
   \caption{Examples of bona-fide data in 6 different sessions, and presentation attacks corresponding to 7 different categories.}
\label{fig:bf_and_pai_samples_small}
\end{figure}

The introduced face PAD system is evaluated on the \textit{Wide Multi-Channel presentation Attack} (\textbf{WMCA}) database (DB), containing a total of \textbf{1679} video files of bona-fide and presentation attacks corresponding to \textbf{72} identities. The temporally synchronized video streams are recorded with two consumer capturing devices, Intel\textsuperscript{\textregistered} RealSense\texttrademark SR300 capturing RGB-NIR-D streams, and Seek Thermal CompactPRO recording the thermal (T) stream. The data collection was split into \textbf{seven} sessions over an interval of \textbf{five} months. Bona-fide samples were recorded in 6 sessions. In each session, a bona-fide and at least two PA performed by the participant were captured. The WMCA DB has more than \textbf{eighty} different presentation attack instruments, which can be groped into \textbf{seven} categories: glasses, fake head, print, replay, rigid mask, flexible/silicon mask, and paper mask. Examples corresponding to these PA categories, as well as the case of bona-fide data for 6 different sessions, are shown in Figure~\ref{fig:bf_and_pai_samples_small}. More detailed information on the number of files for each PA category, and video data technicalities, are summarized in Table~\ref{tab:wmca_db_statistics}. Please refer to~\cite{Anjith_George_IEEET-IFS_2019} for more details on the WMCA database.

\begin{table}[h]
\begin{center}
\footnotesize
\begin{tabular}{|c|c|c|}
\hline
Type          & \#Videos     & Video Details                                                                                                                                        \\ \hline\hline
\textbf{bona-fide}     & 347 (72 IDs) & \multirow{9}{*}{\begin{tabular}[c]{@{}l@{}}Length: \\ 10 seconds\\ ------------------------\\ \#Frames: \\ RGB-NIR-D: 300\\ T: 150\\ ------------------------\\ Format:\\ Uncompressed\end{tabular}} \\ \cline{1-2}
glasses       & 75           &                                                                                                                                                      \\ \cline{1-2}
fake head     & 122          &                                                                                                                                                      \\ \cline{1-2}
print         & 200          &                                                                                                                                                      \\ \cline{1-2}
replay        & 348          &                                                                                                                                                      \\ \cline{1-2}
rigid mask    & 137          &                                                                                                                                                      \\ \cline{1-2}
flexible mask & 379          &                                                                                                                                                      \\ \cline{1-2}
paper mask    & 71           &                                                                                                                                                      \\ \cline{1-2}
\textbf{Total}         & 1679 (5.1 TB)         &                                                                                                                                                      \\ \hline
\end{tabular}
\end{center}
\caption{Main statistics of the WMCA DB.}
\label{tab:wmca_db_statistics}
\end{table}

\section{Experimental evaluation}
\label{sec:experimental_evaluation}

This section covers the details on the evaluation protocols for the WMCA DB, following by the experimental results for baseline MC PAD algorithm, and the results for the proposed PAD setup with different settings. In all experiments, only \textbf{RGB-NIR-D} channels available in WMCA are used, making the proposed PAD system dependent on one capturing device only, specifically Intel\textsuperscript{\textregistered} RealSense\texttrademark SR300.

The WMCA \textbf{evaluation protocol}, namely \textbf{grandtest-10}, follows the legacy evaluation strategy, in this protocol samples of all PAI categories, and bona-fide, are evenly distributed across all subsets: \textit{training}, \textit{development}, and \textit{evaluation}. The identities of bona-fide samples are not intersecting across these subsets. The number \textbf{10}, in the protocol name, stands for the \textit{number of frames} uniformly sampled from each video, thus the total number of biometric samples, using frame-level evaluation strategy, is $1679\cdot10$. Training set is used for training the PAD system. The threshold corresponding to the selected operation points is chosen on the development set, and the system performance is reported on the evaluation set given the threshold.

The \textbf{evaluation metrics} is adopted from the ISO/IEC 30107-3 standard~\cite{iso_pad_standard}, APCER (Attack Presentation Classification Error Rate), and BPCER (Bona-fide Presentation Classification Error Rate). We also adopt two measures, namely BPCER20, and BPCER100, which are the BPCER at APCER $5.0\%$, and $1.0\%$, respectively. Again, the thresholds corresponding to BPCER20, and BPCER100, are selected on the development set, then BPCER, and APCER values are reported on the evaluation set given threshold. In addition to numerical performance, the DET curves are given for all experiments.

\subsection{Results: MC face PAD baselines}
\label{sub:results_baselines}

\begin{table}[]
\begin{center}
\footnotesize
\begin{tabular}{|c|c|c|c|c|}
\hline
Channel  & RGB         & NIR        & D          & Fused           \\ \hline
Method   & IQM+LR      & LBP+LR     & LBP+LR     & LR              \\ \hline\hline
BPCER20  & 77.7        & 9.9        & 13.8       & \textbf{3.0}    \\
APCER    & 13.2        & 7.1        & 9.6        & 10.4            \\ \hline\hline
BPCER100 & 94.6        & 35.6       & 57.5       & \textbf{14.1}   \\
APCER    & 8.5         & 1.9        & 2.0        & 2.9             \\ \hline
\end{tabular}
\end{center}
\caption{Baseline results for \textit{evaluation} set of WMCA, \textbf{grandtest-10} protocol. Thresholds are computed on \textit{development} set.}
\label{tab:results_baselines}
\end{table}

\begin{figure}[t]
\begin{center}
   \includegraphics[width=0.7\linewidth]{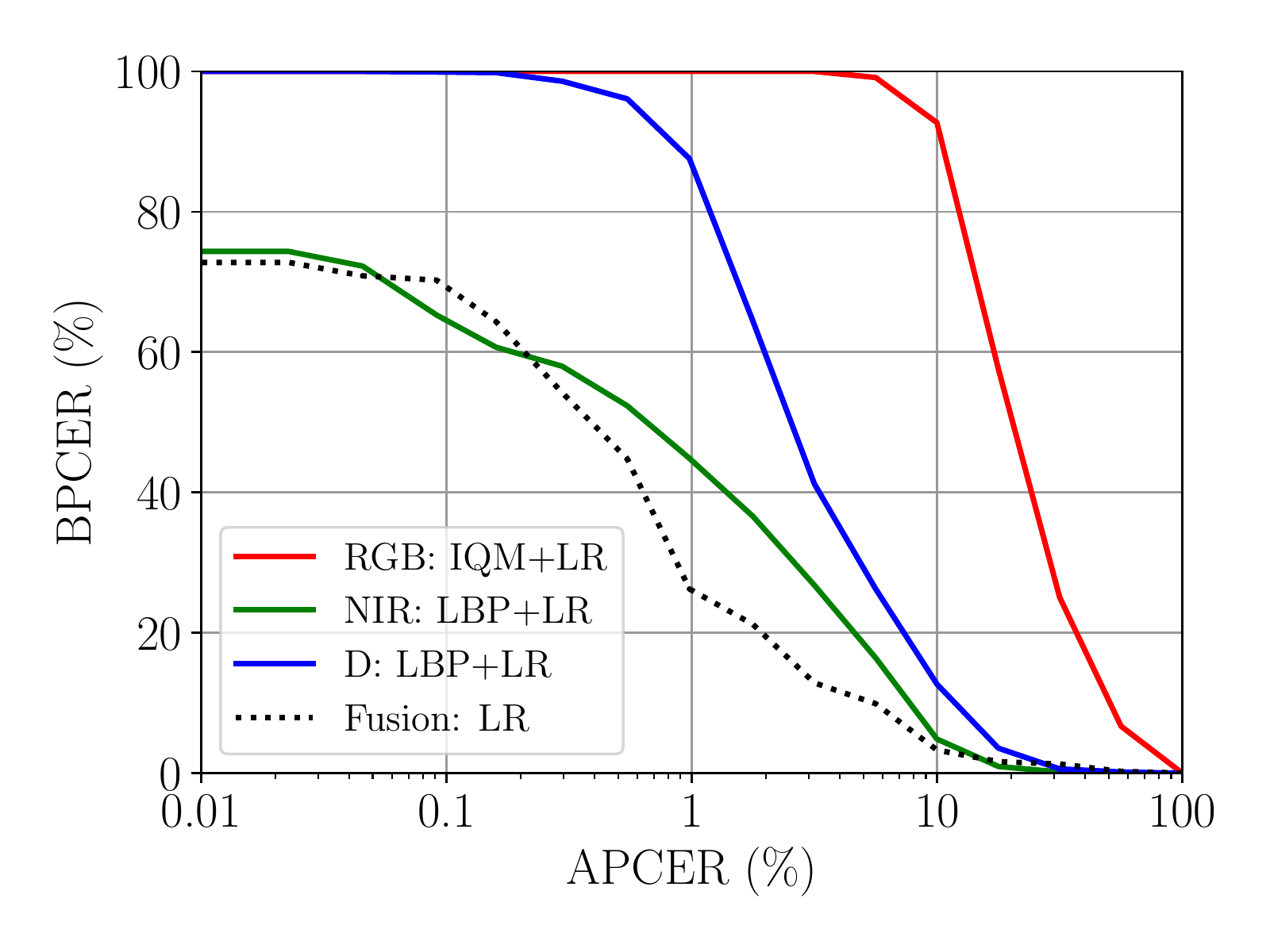}
\end{center}
   \caption{DET curves for baselines. For the \textit{evaluation} set of WMCA, \textbf{grandtest-10} protocol. }
\label{fig:DET_baselines}
\end{figure}

In this subsection a baseline RGB-NIR-D face PAD system is discussed and evaluated, being composed of successful, channel specific, hand-crafted features combined with two-class classifiers. The baseline PAD system is composed of 4 blocks: \textit{preprocessing}, \textit{feature extraction}, \textit{classification}, and \textit{fusion}. The \textbf{preprocessing} stage is similar to the one introduced in Section~\ref{sec:proposed_multichannel_face_pad_approach}, excluding sensor level fusion. In the baseline setup each channel is handled individually, and preprocessed data examples are displayed in Figure~\ref{fig:system_block_diagram}, denoted as RGB, NIR, and D face. The cropped faces are normalized to $64\times64$ pixels, and the frames with faces smaller than $50\times50$ are discarded.

In the \textbf{feature extraction} stage, a grid search across popular hand-crafted features was adopted to identify the best performing ones. The considered features demonstrating reasonable performance in recent literature are Image Quality Measures (IQM)~\cite{Nikisins_anomaly_PAD}, and LBP/MCT histograms~\cite{Nikisins_anomaly_PAD, face-upad}. An IQM is used for the RGB channel, producing feature vectors of 139 quality measures. As a result of a grid-search in the parameter space of spatially enhanced LBP/MCT histograms, an $MCT_{8,1}$ (8 sampling points on a radius of 1) features are chosen for NIR data, and $LBP_{8,1}$ spatially enhanced histograms computed over $2 \times 2$ regions are selected for D frames.

The \textbf{classification} stage deploys a Logistic Regression (LR) for all channels. The features are normalized to zero-mean and unity standard deviation before the training, and the normalization parameters are computed using samples of bona-fide class only. In the prediction stage, a probability of a sample being a PA is computed given pre-trained LR model. A \textbf{fusion} of individual RGB-NIR-D channels is done in the score-level using the same LR-based approach. An LR for the fusion stage is trained using channel scores computed on the \textit{development} set of WMCA.

The results for this sequence of experiments are introduced in Table~\ref{tab:results_baselines}, accumulating BPCER20, BPCER100 and corresponding APCER values on the \textit{evaluation} set of WMCA. Additionally, DET curves are given in Figure~\ref{fig:DET_baselines}. Both from table and DET, one can observe, that MC approach boosts the performance, and legacy RGB baseline operates poorly on the WMCA DB containing a wide range of challenging PAIs.

\subsection{Results: proposed MC face PAD algorithm}
\label{sub:results_proposed_mc_face_pad_algorithm}

\begin{table}[]
\begin{center}
\footnotesize
\begin{tabular}{|c|c|c|c|}
\hline
Method      & \multicolumn{3}{c|}{\textbf{Single AE for entire MC face + MLP}}                                                                                          \\ \hline
AE training & CelebA & \begin{tabular}[c]{@{}c@{}}CelebA+WMCA\\ all encoder layers\end{tabular} & \begin{tabular}[c]{@{}c@{}}CelebA+WMCA\\ 1 encoder layer\end{tabular} \\ \hline\hline
BPCER20     & 3.0    & 1.9                                                                      & 4.1                                                                   \\
APCER       & 4.7    & 4.5                                                                      & 2.8                                                                   \\ \hline
BPCER100    & 59.0   & 51.7                                                                     & 51.6                                                                  \\
APCER       & 0.0    & 0.3                                                                      & 0.0                                                                   \\ \hline\hline
Method      & \multicolumn{3}{c|}{\textbf{9 AE using MC face blocks + MLP}}                                                                                             \\ \hline
AE training & CelebA & \begin{tabular}[c]{@{}c@{}}CelebA+WMCA\\ all encoder layers\end{tabular} & \begin{tabular}[c]{@{}c@{}}CelebA+WMCA\\ 1 encoder layer\end{tabular} \\ \hline\hline
BPCER20     & 1.1    & 5.0                                                                      & 1.5                                                                   \\
APCER       & 5.9    & 3.8                                                                      & 3.1                                                                   \\ \hline
BPCER100    & 11.9   & 12.3                                                                     & \textbf{7.3}                                                          \\
APCER       & 1.1    & 0.7                                                                      & 0.8                                                                   \\ \hline\hline
Method      & \multicolumn{3}{c|}{\textbf{16 AE using MC face blocks + MLP}}                                                                                            \\ \hline
AE training & CelebA & \begin{tabular}[c]{@{}c@{}}CelebA+WMCA\\ all encoder layers\end{tabular} & \begin{tabular}[c]{@{}c@{}}CelebA+WMCA\\ 1 encoder layer\end{tabular} \\ \hline\hline
BPCER20     & 1.7    & 3.0                                                                      & \textbf{1.0}                                                          \\
APCER       & 3.1    & 3.6                                                                      & 3.5                                                                   \\ \hline
BPCER100    & 10.7   & 16.4                                                                     & 20.4                                                                  \\
APCER       & 0.7    & 0.6                                                                      & 0.2                                                                   \\ \hline
\end{tabular}
\end{center}
\caption{BPCER20, BPCER100, and corresponding APCER in $\%$, for the proposed MC face PAD system, reported for \textit{evaluation} set of WMCA, \textbf{grandtest-10} protocol. Experiments for different regioning of facial images, and different AE training strategies. Thresholds are computed on \textit{development} set.}
\label{tab:results_mc_autoencoders}
\end{table}

\begin{figure}[t]
\begin{center}
   \includegraphics[width=0.7\linewidth]{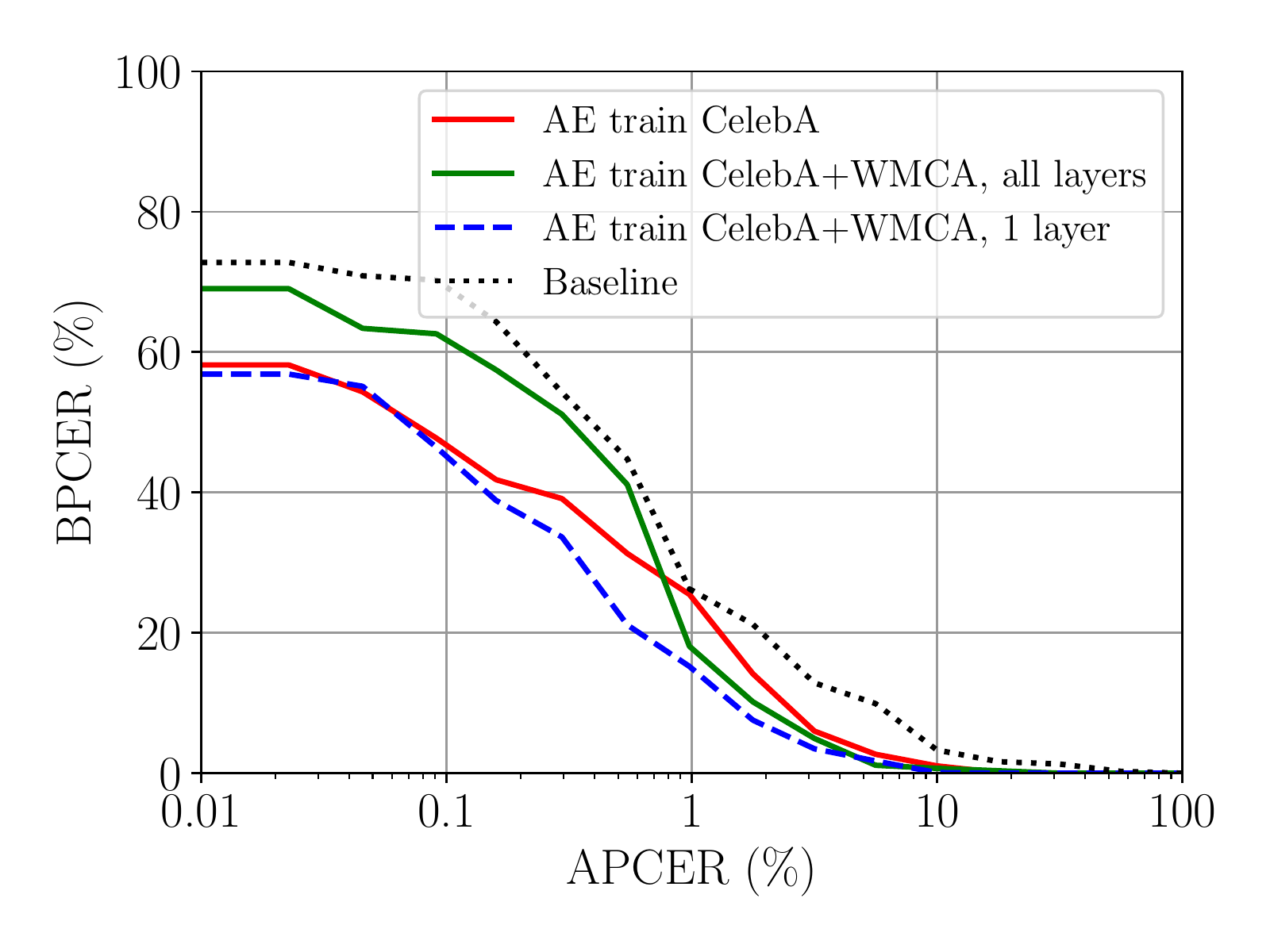}
\end{center}
   \caption{DET curves for PAD system using AE for the \textit{entire} MC face, and MLP classifier. For the \textit{evaluation} set of WMCA, \textbf{grandtest-10} protocol.}
\label{fig:DET_entire_face_AE}
\end{figure}

\begin{figure}[t]
\begin{center}
   \includegraphics[width=0.7\linewidth]{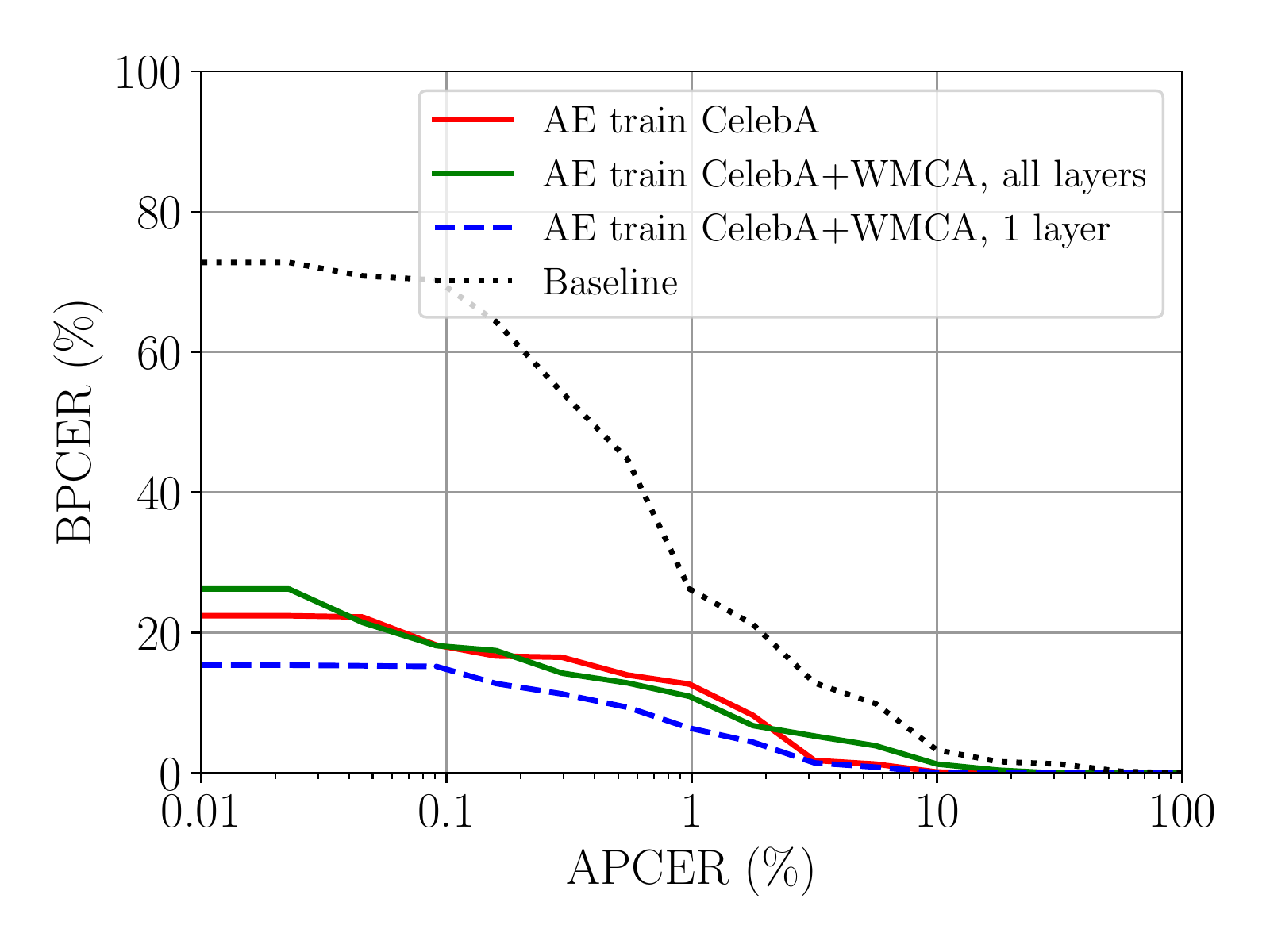}
\end{center}
   \caption{DET curves for PAD system using 9 AE for MC facial blocks, and MLP classifier. For the \textit{evaluation} set of WMCA, \textbf{grandtest-10} protocol.}
\label{fig:DET_9_AE}
\end{figure}

\begin{figure}[t]
\begin{center}
   \includegraphics[width=0.7\linewidth]{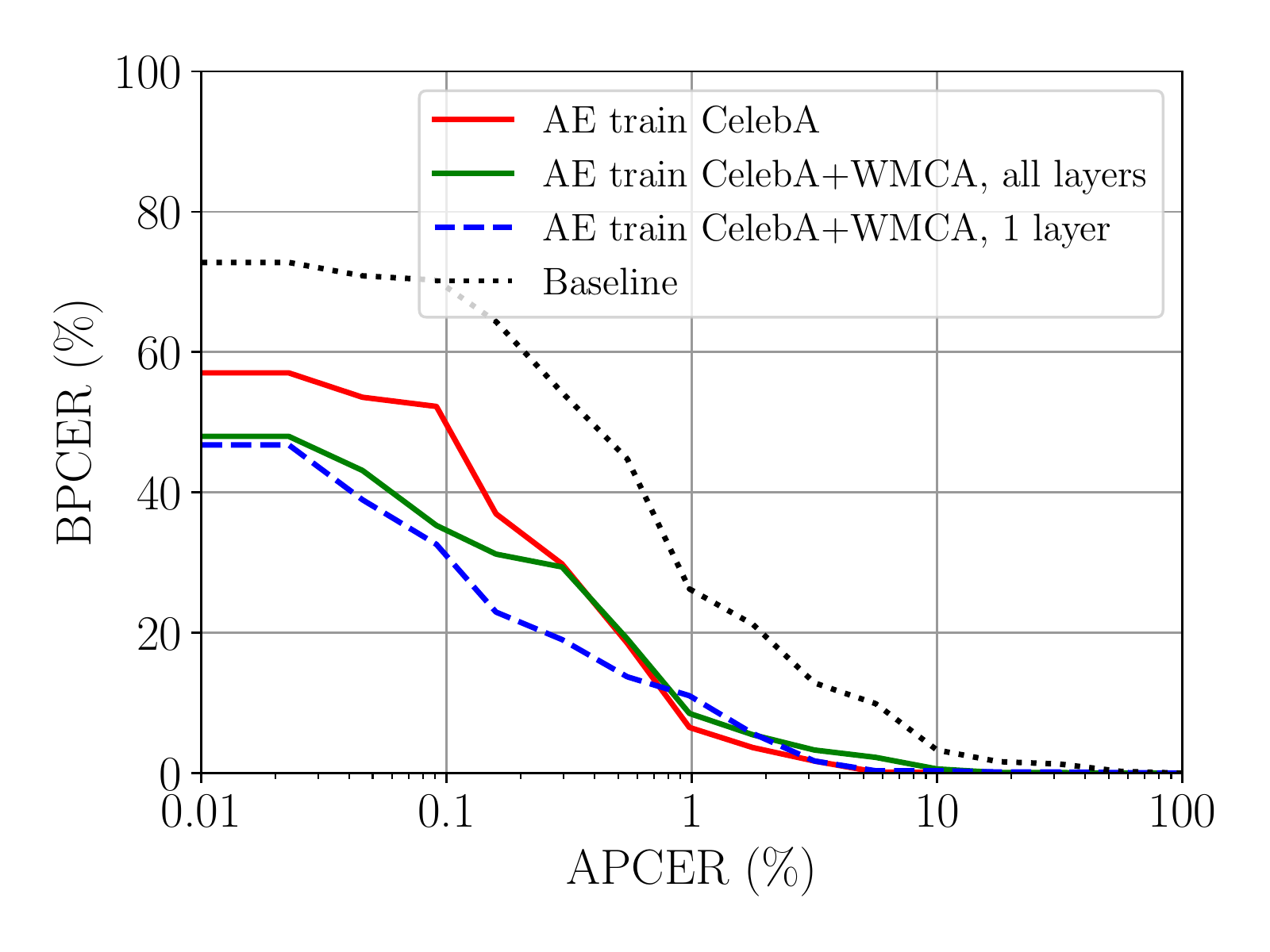}
\end{center}
   \caption{DET curves for PAD system using 16 AE for MC facial blocks, and MLP classifier. For the \textit{evaluation} set of WMCA, \textbf{grandtest-10} protocol.}
\label{fig:DET_16_AE}
\end{figure}

In this subsection a proposed CNN-based BW-NIR-D face PAD system is evaluated under different training scenarios and parametrization. Training of the network incorporates both RGB data from CelebA, and BW-NIR-D data from WMCA databases. To be suitable for training the biometric recordings are first preprocessed, as discussed in Section~\ref{sec:proposed_multichannel_face_pad_approach}, producing facial images of the size $128_{(pixels)}\times128_{(pixels)}\times3_{(chanels)}$ in both RGB and MC cases. The preprocessed RGB data from CelebA database undergoes additional quality assessment procedure, before being used for training. More specifically, an eye detection algorithm is applied to face images, assuring the deviation of eye coordinates from expected positions is not significant. The Haar-based eye detector from OpenCV is integrated for this purpose. This is done to exclude images with occlusions, \eg sun-glasses, which can be considered as PAI, rather than bona-fide. The resulting amount of \textit{bona-fide} RGB and MC training images is 42499, and 1240 respectively. An example of preprocessed/training data is displayed in Figure~\ref{fig:system_block_diagram}, denoted as RGB face and MC face.

\textbf{Three training strategies} are explored in current experiments: $N$ autoencoders (AEs) are trained using only RGB CelebA facial regions (\textit{no domain adaptation}); $N$ AEs are trained using RGB CelebA facial regions, and fully fine-tuned using MC facial regions (RGB-to-MC domain adaptation, \textit{full fine-tuning}); $N$ AEs are trained using RGB CelebA facial regions, and only first and 2 last layers are fine-tuned using MC facial regions (RGB-to-MC domain adaptation, \textit{partial fine-tuning}). As discussed in Section~\ref{sec:proposed_multichannel_face_pad_approach}, training of the proposed network is associated to two components: a set of convolutional AEs, and an
MLP. Here, training of an MLP remains the same, for all three strategies of autoencoders training. An MLP is trained using latent features (concatenation of latent vectors of all AEs) of both bona-fide and attack classes present in the training set of the WMCA database. In each experiment, an MLP is re-trained 10 times, initializing it differently. The best MLP model is selected cross-validating them on the \textit{development} set of the WMCA.
A BCE loss is used in both training and cross-validation of an MLP. AEs are trained with MSE loss in the unsupervised manner, running training and fine-tuning for 50 epoch each. The detailed parametrization of autoencoders and MLP is displayed in Figure~\ref{fig:system_block_diagram}.

Additionally, \textbf{three face regioning approaches} are observed for each training strategy discussed above. In the \textit{first} sequence of experiments, an autoencoder is trained using entire face region, $N=1$. \textit{Second} and \textit{third} approaches assume splitting of the facial region into $N=9$, and $N=16$ regions respectively, training an individual AE for each region. The dimensionality of RGB/MC facial blocks is $64_{(pixels)}\times64_{(pixels)}\times3_{(chanels)}$ for $N=9$, and is $32_{(pixels)}\times32_{(pixels)}\times3_{(chanels)}$ for $N=16$, with a patching stride of 32 pixels in both cases. The dimensionalities of latent feature spaces are $1296$ for $N=1$, $3600$ for $N=9$, and $2304$ for $N=16$.

The results for this sequence of experiments are introduced in Table~\ref{tab:results_mc_autoencoders}, summarizing BPCER20, BPCER100 and corresponding APCER values on the \textit{evaluation} set of WMCA. The DET curves are given in Figure~\ref{fig:DET_entire_face_AE} for $N=1$, Figure~\ref{fig:DET_9_AE} for $N=9$, and Figure~\ref{fig:DET_16_AE} for $N=16$.
From Table~\ref{tab:results_mc_autoencoders} one can see, that top BPCER20 and BPCER100 values are observed in the case of AEs training incorporating \textit{partial fine-tuning}, and with \textit{face regioning}. The same trend can be observed in the DET curves, with clearly \textbf{best performing PAD system} based on $N=9$ autoencoders, which are pre-trained using RGB CelebA facial regions, followed by partial fine-tuning on MC data from WMCA, see Figure~\ref{fig:DET_9_AE}. Also, latent features learned combining bona-fide training samples from RGB and MC domains, demonstrate the superior discriminative capacity, as opposed to the MC baselines using hand-crafted features, especially in the range of low APCER values, Figure~\ref{fig:DET_9_AE}.

Interestingly, that \textit{full AEs fine-tuning} using MC data doesn't boost, or even degrades, the overall performance of the PAD system in all face regioning approaches, Figures~\ref{fig:DET_entire_face_AE} -~\ref{fig:DET_16_AE}. While the proposed \textit{partial fine-tuning} has a stable positive impact, as opposed to the AEs trained using CelebA data only. It is worth mentioning, that we have also tested other RGB-to-MC domain adaptation strategies, fine-tuning more than just one layer of encoders, but it doesn't improve the performance further.

\subsection{Results: proposed face PAD in RGB mode}
\label{sub:results_features_of_9_facial_rgb_blocks}

\begin{table}[]
\begin{center}
\footnotesize
\begin{tabular}{|c|c|c|}
\hline
Channel  & \multicolumn{2}{c|}{RGB}                                                        \\ \hline
Method   & IQM+LR & \begin{tabular}[c]{@{}c@{}}9 AE trained on\\ CelebA + MLP\end{tabular} \\ \hline\hline
BPCER20  & 77.7      & \textbf{10.5}                                                       \\
APCER    & 13.2      & 17.3                                                                \\ \hline\hline
BPCER100 & 94.6      & \textbf{29.2}                                                       \\
APCER    & 8.5       & 7.8                                                                 \\ \hline
\end{tabular}
\end{center}
\caption{BPCER20, BPCER100, and corresponding APCER in $\%$, using RGB channel only for \textit{evaluation} set of WMCA, \textbf{grandtest-10} protocol. Thresholds are computed on \textit{development} set.}
\label{tab:results_rgb_autoencoders}
\end{table}

\begin{figure}[t]
\begin{center}
   \includegraphics[width=0.7\linewidth]{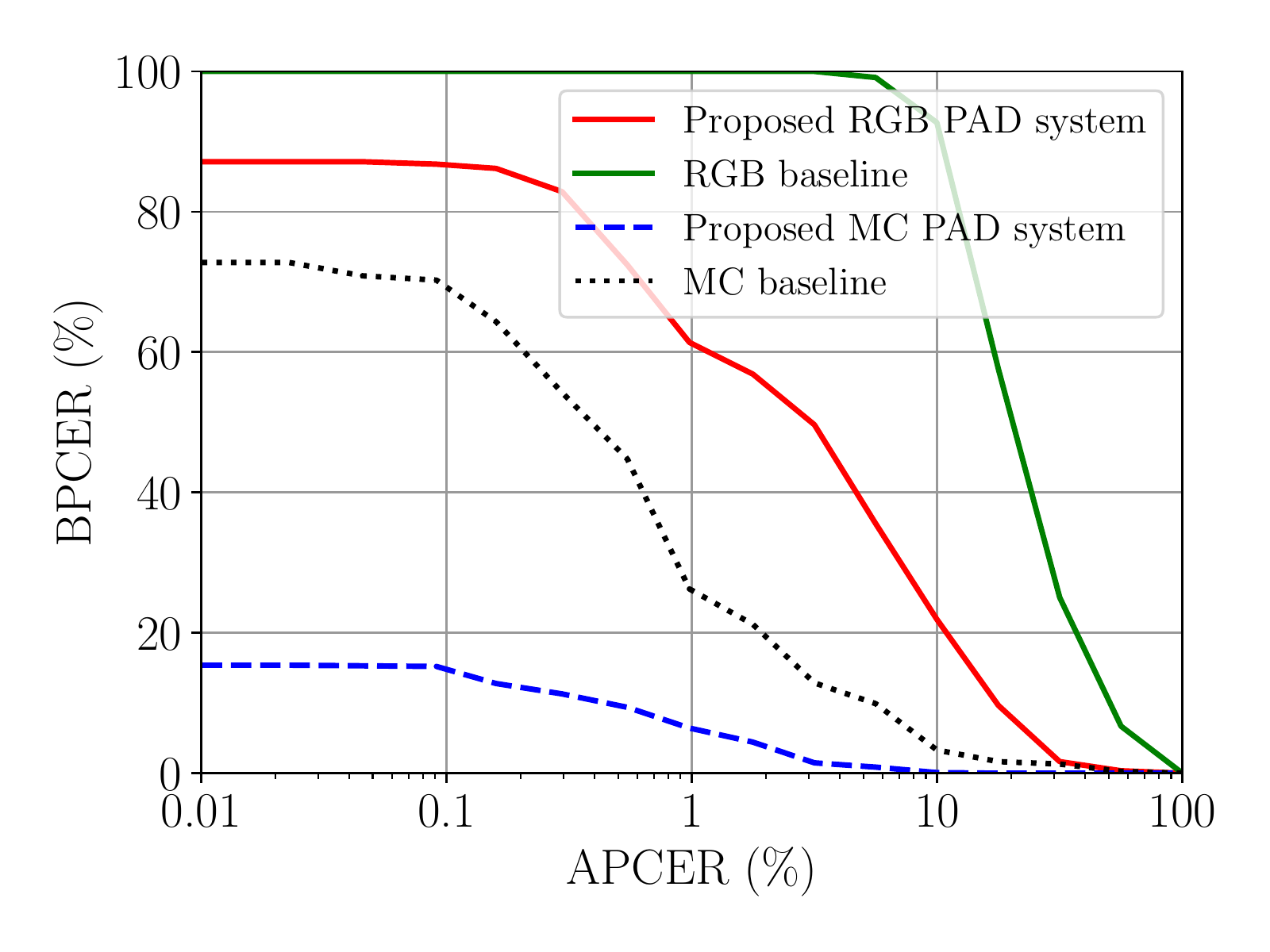}
\end{center}
   \caption{DET curves for proposed RGB PAD system (using 9 AE for RGB facial blocks, and MLP classifier) and RGB baseline, vs. proposed MC PAD approach and MC baseline. For the \textit{evaluation} set of WMCA, \textbf{grandtest-10} protocol.}
\label{fig:DET_9_RGB_AEs}
\end{figure}

In the final sequence of experiments, the introduced face PAD approach is tested in RGB mode, meaning that AEs are trained using RGB CelebA facial regions only, and an MLP is trained using latent encodings obtained from RGB channel of WMCA. No fine-tuning of AEs is involved. The input data in the experiments is RGB channel, present in the WMCA DB. Other parameters of the system, \eg preprocessing and face regioning ($N=9$), are the same as in the \textit{best performing} system of subsection~\ref{sub:results_proposed_mc_face_pad_algorithm}.
Following the results summarized in Table~\ref{tab:results_rgb_autoencoders} and Figure~\ref{fig:DET_9_RGB_AEs}, offered RGB CNN-based system outperforms the IQM-based RGB baseline by a large margin.
However, an auxiliary discriminative information introduced by MC approach seems to have a higher impact on the performance, rather than using advanced PAD methods in RGB domain only, see Figure~\ref{fig:DET_9_RGB_AEs}.

\section{Conclusion}
\label{sec:conclusion}

Current paper suggests a multi-channel face anti-spoofing solution, being motivated by the need of high security PAD systems, capable of dealing with challenging attacks, such as 3D and partial PAIs. The introduced approach is using a consumer grade imaging sensors, capturing BW-NIR-D streams, in combination with a deep-learning based PAD algorithm. In the experimental section we demonstrate, that heavily studied RGB-based legacy systems are inefficient in detecting a wide range of challenging PAIs present in the WMCA database. In contrast, our proposed solution gives very promising results, especially in the range of low APCER values, which is critical for high security applications, see Figure~\ref{fig:DET_9_RGB_AEs}.

Current paper introduces a number of novelties.
\textit{First}, a CNN-based MC face PAD algorithm, which is decomposed into a set of encoders, processing individual MC facial regions, and an MLP, categorizing latent encodings into real or attacks classes.
\textit{Second}, CNN decomposition allows us to introduce a special training procedure, transferring the knowledge of facial appearance from RGB to multi-channel domain. Domain adaptation is done via autoencoders, which are first pre-trained on a large set of RGB facial data from CelebA, and are then \textit{partially fine-tuned} on the BW-NIR-D data from WMCA DB. Here partial fine-tuning means training of the first layers of encoders, and last 2 layers of decoders (for symmetry). The intuition behind this proposition, is that only low level features are domain dependent, while deeper features/layers of AEs are domain independent mostly preserving \textit{structural information}
of the face. \textit{Full fine-tuning} of the AEs will try to learn this \textit{structural information} from a training set of WMCA database, which has a significantly smaller number of \textit{bona-fide} training samples, than CelebA DB, leading to over-fitting of the system.
The effectiveness of the proposed RGB-to-MC domain adaptation, via partial fine-tuning, is proved experimentally. \textit{Third}, we demonstrate, that learning the features of individual facial regions, is more discriminative than the features learned from an entire face.

To the best of our knowledge, this work is one of the first attempts applying deep-learning technologies, and domain adaptation, to the task of multi-channel face PAD, giving promising results and motivating to enhance the research in this direction.

\section*{Acknowledgment}

This research is based upon work supported by the Office of the Director of National Intelligence (ODNI), Intelligence Advanced Research Projects Activity (IARPA), via IARPA R\&D Contract No. 2017-17020200005. The views and conclusions contained herein are those of the authors and should not be interpreted as necessarily representing the official policies or endorsements, either expressed or implied, of the ODNI, IARPA, or the U.S. Government. The U.S. Government is authorized to reproduce and distribute reprints for Governmental purposes notwithstanding any copyright annotation thereon.

{\small
\bibliographystyle{ieee}
\bibliography{egbib}

\begin{thebibliography}{10}\itemsep=-1pt

\bibitem{iso_pad_standard}
{\em Information technology - Biometric presentation attack detection - Part 3:
  Testing and reporting}, 2017.

\bibitem{Kittler_anomaly_PAD}
S.~R. Arashloo and J.~Kittler.
\newblock An anomaly detection approach to face spoofing detection: A new
  formulation and evaluation protocol.
\newblock In {\em 2017 IEEE International Joint Conference on Biometrics
  (IJCB)}, pages 80--89, Oct 2017.

\bibitem{8053524}
S.~Bhattacharjee and S.~Marcel.
\newblock What you can't see can help you - extended-range imaging for 3d-mask
  presentation attack detection.
\newblock In {\em 2017 International Conference of the Biometrics Special
  Interest Group}, pages 1--7, Sept 2017.

\bibitem{Bhattacharjee_BTAS2018_2018}
S.~Bhattacharjee, A.~Mohammadi, and S.~Marcel.
\newblock Spoofing deep face recognition with custom silicone masks.
\newblock In {\em Proceedings of BTAS2018}, Oct. 2018.

\bibitem{DBLP:journals/corr/abs-1802-05798}
A.~Bhattad, J.~Rock, and D.~A. Forsyth.
\newblock Detecting anomalous faces with 'no peeking' autoencoders.
\newblock {\em CoRR}, abs/1802.05798, 2018.

\bibitem{OULU_NPU_2017}
Z.~Boulkenafet, J.~Komulainen, L.~Li, X.~Feng, and A.~Hadid.
\newblock {OULU-NPU}: A mobile face presentation attack database with
  real-world variations.
\newblock May 2017.

\bibitem{Chingovska_BIOSIG-2012}
I.~Chingovska, A.~Anjos, and S.~Marcel.
\newblock On the effectiveness of local binary patterns in face anti-spoofing.
\newblock 2012.

\bibitem{Costa-Pazo_BIOSIG2016_2016}
A.~Costa-Pazo, S.~Bhattacharjee, E.~Vazquez-Fernandez, and S.~Marcel.
\newblock The replay-mobile face presentation-attack database.
\newblock In {\em Proceedings of the International Conference on Biometrics
  Special Interests Group (BioSIG)}, Sept. 2016.

\bibitem{deFreitasPereira_IEEET-IFS_2019}
T.~de~Freitas~Pereira, A.~Anjos, and S.~Marcel.
\newblock Heterogeneous face recognition using domain specific units.
\newblock {\em IEEE Transactions on Information Forensics and Security},
  page~13, Feb. 2019.

\bibitem{DBLP:journals/corr/abs-1806-07492}
G.~B. de~Souza, J.~P. Papa, and A.~N. Marana.
\newblock On the learning of deep local features for robust face spoofing
  detection.
\newblock {\em CoRR}, abs/1806.07492, 2018.

\bibitem{IJST120575}
S.~N. Garg, R.~Vig, and S.~Gupta.
\newblock A survey on different levels of fusion in multimodal biometrics.
\newblock {\em Indian Journal of Science and Technology}, 10(44), 2017.

\bibitem{Anjith_George_IEEET-IFS_2019}
A.~George, Z.~Mostaani, D.~Geissenbuhler, O.~Nikisins, A.~Anjos, and S.~Marcel.
\newblock Biometric face presentation attack detection with multi-channel
  convolutional neural network.
\newblock {\em IEEE Transactions on Information Forensics and Security (under
  review)}, 2019.

\bibitem{7821013}
L.~Li, X.~Feng, Z.~Boulkenafet, Z.~Xia, M.~Li, and A.~Hadid.
\newblock An original face anti-spoofing approach using partial convolutional
  neural network.
\newblock In {\em 2016 Sixth International Conference on Image Processing
  Theory, Tools and Applications (IPTA)}, pages 1--6, Dec 2016.

\bibitem{siw_db}
Y.~Liu*, A.~Jourabloo*, and X.~Liu.
\newblock Learning deep models for face anti-spoofing: Binary or auxiliary
  supervision.
\newblock In {\em In Proceeding of IEEE Computer Vision and Pattern
  Recognition}, Salt Lake City, UT, June 2018.

\bibitem{DBLP:conf/cvpr/LiuJ018}
Y.~Liu, A.~Jourabloo, and X.~Liu.
\newblock Learning deep models for face anti-spoofing: Binary or auxiliary
  supervision.
\newblock In {\em 2018 {IEEE} Conference on Computer Vision and Pattern
  Recognition, {CVPR} 2018, Salt Lake City, UT, USA, June 18-22, 2018}, pages
  389--398, 2018.

\bibitem{liu2015faceattributes}
Z.~Liu, P.~Luo, X.~Wang, and X.~Tang.
\newblock Deep learning face attributes in the wild.
\newblock In {\em Proceedings of International Conference on Computer Vision
  (ICCV)}, 2015.

\bibitem{Mohammadi_IETBIOMETRICS_2017}
A.~Mohammadi, S.~Bhattacharjee, and S.~Marcel.
\newblock Deeply vulnerable -- a study of the robustness of face recognition to
  presentation attacks.
\newblock {\em IET (The Institution of Engineering and Technology) --
  Biometrics}, pages 1--13, 2017.
\newblock Accepted on 29-Sept-2017.

\bibitem{Nikisins_anomaly_PAD}
O.~Nikisins, A.~Mohammadi, A.~Anjos, and S.~Marcel.
\newblock On effectiveness of anomaly detection approaches against unseen
  presentation attacks in face anti-spoofing.
\newblock In {\em 2018 International Conference on Biometrics (ICB)}, 2018.

\bibitem{Parkhi15}
O.~M. Parkhi, A.~Vedaldi, and A.~Zisserman.
\newblock Deep face recognition.
\newblock In {\em British Machine Vision Conference}.

\bibitem{8009749}
R.~Raghavendra, K.~B. Raja, S.~Venkatesh, and C.~Busch.
\newblock Extended multispectral face presentation attack detection: An
  approach based on fusing information from individual spectral bands.
\newblock In {\em 2017 20th International Conference on Information Fusion
  (Fusion)}, pages 1--6, July 2017.

\bibitem{Ramachandra:2017:PAD:3058791.3038924}
R.~Ramachandra and C.~Busch.
\newblock Presentation attack detection methods for face recognition systems: A
  comprehensive survey.
\newblock {\em ACM Comput. Surv.}, 50(1):8:1--8:37, Mar. 2017.

\bibitem{7746050}
M.~O. Simón, C.~Corneanu, K.~Nasrollahi, O.~Nikisins, S.~Escalera, Y.~Sun,
  H.~Li, Z.~Sun, T.~B. Moeslund, and M.~Greitans.
\newblock Improved rgb-d-t based face recognition.
\newblock {\em IET Biometrics}, 5(4):297--303, 2016.

\bibitem{DBLP:journals/js/SteinerSKJ16}
H.~Steiner, S.~Sporrer, A.~Kolb, and N.~Jung.
\newblock Design of an active multispectral {SWIR} camera system for skin
  detection and face verification.
\newblock {\em J. Sensors}, 2016:9682453:1--9682453:16, 2016.

\bibitem{WenTIFS15}
D.~Wen, H.~Han, and A.~Jain.
\newblock { Face Spoof Detection with Image Distortion Analysis}.
\newblock {\em IEEE Trans. Information Forensic and Security}, 10(4):746--761,
  April 2015.

\bibitem{face-upad}
F.~Xiong and W.~Abdalmageed.
\newblock Unknown presentation attack detection with face rgb images.
\newblock In {\em IEEE International Conference on Biometrics: Theory,
  Applications, and Systems}, 2018.

\bibitem{7553523}
K.~Zhang, Z.~Zhang, Z.~Li, and Y.~Qiao.
\newblock Joint face detection and alignment using multitask cascaded
  convolutional networks.
\newblock {\em IEEE Signal Processing Letters}, 23(10), 2016.

\bibitem{6199754}
Z.~Zhang, J.~Yan, S.~Liu, Z.~Lei, D.~Yi, and S.~Z. Li.
\newblock A face antispoofing database with diverse attacks.
\newblock In {\em 2012 5th IAPR International Conference on Biometrics (ICB)},
  pages 26--31, March 2012.

\end{thebibliography}
}

\end{document}